\documentclass[nohyperref]{article}

\usepackage{microtype}
\usepackage{graphicx}
\usepackage{subfigure}
\usepackage{booktabs} 
\usepackage{multirow}

\usepackage{hyperref}

\usepackage[accepted]{icml2022}

\usepackage{amsmath}
\usepackage{amssymb}
\usepackage{mathtools}
\usepackage{amsthm}
\usepackage{graphicx}
\usepackage{float}
\usepackage{subfigure}
\usepackage{color}
\usepackage{enumitem}

\usepackage[capitalize,noabbrev]{cleveref}

\theoremstyle{plain}
\newtheorem{theorem}{Theorem}[section]

\theoremstyle{definition}
\newtheorem{definition}[theorem]{Definition}

\theoremstyle{remark}

\usepackage[textsize=tiny]{todonotes}

\icmltitlerunning{Neural Collapse Inspired Federated Learning with Non-iid Data}

\begin{document}

\twocolumn[
\icmltitle{Neural Collapse Inspired Federated Learning with Non-iid Data}



\icmlsetsymbol{equal}{*}

\begin{icmlauthorlist}
\icmlauthor{Chenxi Huang}{equal,sch}
\icmlauthor{Liang Xie}{equal,sch}
\icmlauthor{Yibo Yang}{other}
\icmlauthor{Wenxiao Wang}{sch}
\icmlauthor{BinBin Lin}{sch}
\icmlauthor{Deng Cai}{sch}
\end{icmlauthorlist}

\icmlaffiliation{sch}{They are with the State Key Laboratory of CAD\&CG, College of Computer Science, Zhejiang University, Hangzhou, Zhejiang 310058, China.}
\icmlaffiliation{other}{He is the researcher at JD Explore Academy}

\icmlcorrespondingauthor{Chenxi Huang}{hcx\_98@zju.edu.cn}

\icmlkeywords{Federated Learning, Neural Collapse, non-IID}

\vskip 0.3in
]



\printAffiliationsAndNotice{\icmlEqualContribution} 

\begin{abstract}
One of the challenges in federated learning is the non-independent and identically distributed (non-iid) characteristics between heterogeneous devices, which cause significant differences in local updates and affect the performance of the central server.
Although many studies have been proposed to address this challenge, they only focus on local training and aggregation processes to smooth the changes and fail to achieve high performance with deep learning models.
Inspired by the phenomenon of neural collapse, we force each client to be optimized toward an optimal global structure for classification.
Specifically, we initialize it as a random simplex Equiangular Tight Frame (ETF) and fix it as the unit optimization target of all clients during the local updating.
After guaranteeing all clients are learning to converge to the global optimum, we propose to add a global memory vector for each category to remedy the parameter fluctuation caused by the bias of the intra-class condition distribution among clients. 
Our experimental results show that our method can improve the performance with faster convergence speed on different-size datasets.\looseness=-1
\end{abstract}

\section{Introduction}
\label{intro.}
Due to the data privacy or the transmission costs, data from independent individuals cannot be shared on the same global server in various real-world applications. 
Federated learning~\cite{FL} is a concept that leverages data spreading across many devices to learn specific tasks distributively without recourse to data sharing.
The fundamental federated learning problem can be cast as an empirical minimization of a global loss objective but learns from the local private datasets.
Specifically, federated learning requires multiple rounds of local and global update processes to aggregate clients with privacy requirements. 
Some clients are randomly selected and initialized by the global model from the central server. 
They update models independently with their local private data. 
After the local training is completed, the central server updates the global model by aggregating local models. 
The above two processes are repeated until convergence. 
Only model parameters are transmitted between clients and the global server in federated learning, ensuring basic privacy protection requirements.

FedAvg~\cite{FedAvg} is a widely used method for federated learning, which can significantly reduce the communication costs.
Despite the solid empirical performance of FedAvg in Independent Identically Distributed (iid) settings, its performance declines in non-iid settings~\cite{FedNoniid}.
The non-iid challenges come from two aspects.
First, some clients only can observe part of the entire categories in the training set. 
It causes the local learning program only seeks to minimize the objective function under the observed categories.
And the knowledge of missing categories from the central server will be forgotten in the local updating.
Second, even for the observed categories, the data distributions of different clients are not identical. 
The intra-class non-iid distributions make parameters of the feature extractor update in different directions, which causes severe fluctuation and inefficiency during learning.
Several variants of FedAvg~\cite{FedNoniid,FedavgNoniid,FedRS} have been proposed to handle non-iid settings via adding proximal terms to the objective function or using smoother aggregation. 
However, they do not fundamentally solve the problem caused by the non-iid distribution and have no significant improvement in performance.\looseness=-1

The core problem caused by the non-iid challenges lies in the lack of a consistent optimization target across all clients.
Therefore, we must create a suitable optimization target for all clients and make them learn in the same direction.
Neural collapse~\cite{NeuralCollapse} finds that the optimal solution in the balanced dataset under cross-entropy loss is a simplex Equiangular Tight Frame (ETF) geometric structure for classification tasks.
This structure enjoys the most prominent possible equiangular separation of all classes. 
Inspired by this property, in this paper, we propose to directly initialize a global classifier as a random simplex ETF geometric structure and fix it during training, i.e., only the backbone can be learned.
Before a client updates locally, it will be initialized by the model from the central server.
Even if the data of \textbf{different clients} is non-iid, the structure of the classifier will not be damaged, and the final classifier of the central server will present an equal pair-wise margin for all classes.\looseness=-1

Although the feature is optimized in the direction of the pre-assigned target (a simplex ETF geometric structure), the differences in intra-class condition distribution will still lead to fluctuations in backbone learning since different clients have different prior probabilities.
Therefore, after ensuring that all clients are optimized toward the global optimum, we propose adding a global memory vector for each category to compensate for the fluctuations caused by the deviation of the intra-class conditional distribution between clients.\looseness=-1

The main contributions of this study can be listed as follows:
\begin{itemize}
    \item Inspired by the phenomenon of neural collapse, we propose a solution for the non-iid problem in federated learning by directly adding a consistent optimization target across local clients. 
    \item To mitigate the fluctuation caused by the intra-class inconsistency of the conditional distribution among clients, we propose maintaining a global memory vector for each category to correct the bias in the backbone training.\looseness=-1
    \item We demonstrate the generality and the solid experimental performance of our method, significantly improving upon the commonly used methods on multiple federated learning benchmarks. 
\end{itemize}

\section{Related Works}
\label{rel.}
\subsection{Federated Learning with Non-iid Data}
FedAvg~\cite{FedAvg}, as the most standard FL algorithm, proposes using a large number of local SGD steps per round. 
In each round of FedAvg, the updated local models of the clients are transferred to the server, which further aggregates the local models to update the global model. 
However, the performance degrades in non-iid scenes~\cite{FedavgNoniid}.
Several varieties of FedAvg have been proposed to handle non-iid settings.
Those studies can be divided into two categories: improving on local training~\cite{FedProx,scaffold,FedDyn,FedBN,ImbalanceFed} and improving on aggregation~\cite{FedAvgM,FedNova,FedMA}.

As for studies on improving local training, FedProx~\cite{FedProx} introduces a proximal term as a dynamic regularizer into the objective during local training. 
The proximal term is computed based on an l2-norm distance between the current round global model and the local model, penalizing when the updates are far away from the server model.  
Thus, the local model update is limited by the proximal term during the local training.
SCAFFOLD~\cite{scaffold} corrects the local updates by adding control variates. 
Like the model of the server, the global control variates are also updated by all clients during aggregation. 
The difference between the local and global control variate is used to correct the gradients in local training.
However, the experiments in MOON~\cite{Moon} show that they fail to achieve good performance on image datasets with deep learning models, which can be as bad as FedAvg.
FedRS~\cite{FedRS} points out that the existing Softmax+Cross Entropy will cause the classification weight of the missing category to be updated in the wrong direction, so Restricted Softmax is proposed. 
However, special processing is only made at the last layer of the neural network, which is more suitable for shallow neural networks. 

As for studies on improving aggregation, SCAFFOLD~\cite{scaffold} sets momentum updates in the central server, achieving stable performance. 
FedMA~\cite{FedMA} utilizes Bayesian non-parametric methods to match and average weights in a layer-wise manner. 
FedNova~\cite{FedNova} normalizes the local updates before averaging. 
FedMMD~\cite{FedMMD} aims to mitigate the feature discrepancy between local and global models.
These methods increase the stability and performance of the central server by avoiding excessive updates.
However, the above methods only smoother the learning process but do not fundamentally solve the problem of non-iid.\looseness=-1

Recently, SphereFed~\cite{SphereFed} has proposed to solve the problem from another perspective, limiting the learning representation.
It fixes the weights of the classifier to a unit hypersphere, which clients share. 
Similarly, we restrict the learning of all clients, but we directly take an optimal structure as the unified optimization target.

\subsection{Neural Collapse}
In~\cite{NeuralCollapse}, the neural collapse was observed at the last stage of training on the balanced dataset.
Moreover, it proves that the optimal classifier vectors under MSE loss will converge to neural collapse.
Specifically, neural collapse is a phenomenon, that is, the average of the intra-class feature and the classifier weight vectors will converge to an ETF geometric structure if the distance between the feature vectors and the classifier weights is measured by Euclidean distance.
\cite{GeometricAnalysis,CrossEntropy,UnconstrainedLayer,Dissecting,Unconstrainedfeatures,GeometricUnconstrained,MSELoss} analyze the neural collapse phenomenon under a balanced data distribution. 
They only take the last layer of features and the classifier as independent variables. 
They prove that neural collapse would occur under the CE loss with appropriate constraints or regularization. 
However, for most real-world data, the class distribution is usually imbalanced.
\cite{yang2022we} shows theoretically that neural collapse also can occur in imbalanced training with a fixed ETF classifier and propose a new loss function with better convergence properties and imbalanced classification performance than the common CE loss.
However, it is not guaranteed that a fixed length of classifier weight does not harm representation learning.
Therefore, \cite{xie2022neural} proposes Attraction-Repulsion-Balanced Loss (ARB-Loss) from the neural collapse view to solve long-tailed tasks. \looseness=-1

As data distributions of each client in federated learning may differ, the classifiers of clients are trained in different directions. 
Therefore, we can not be aggregated with the local models directly.
Instead, we force each client to be optimized to the optimal global structure for classification.
Inspired by neural collapse, we directly fixed the local model of clients as the structure of the optimal solution, i.e., a random simplex ETF geometric structure.

\begin{figure}[tbp]
\centering
\includegraphics[width=0.96\linewidth]{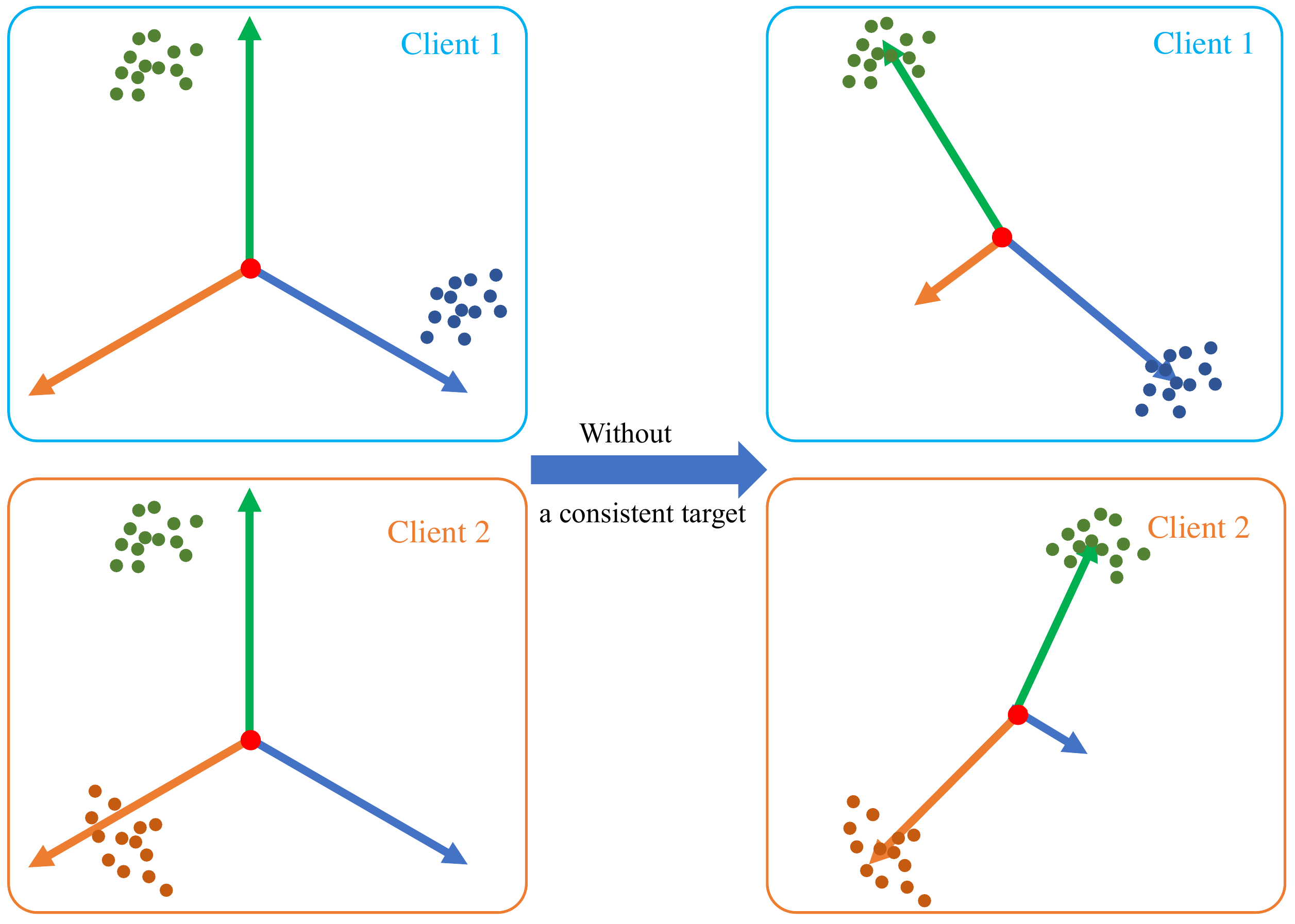}
\caption{The phenomenon of non-iid that $\mathcal{P}_k(y)$ are different. Different colors mean different categories. Points represent samples and arrows represent the weights of the local classifier.}
\label{Fig:py}
\end{figure}

\begin{figure}[tb]
\centering
    \includegraphics[width=0.96\linewidth]{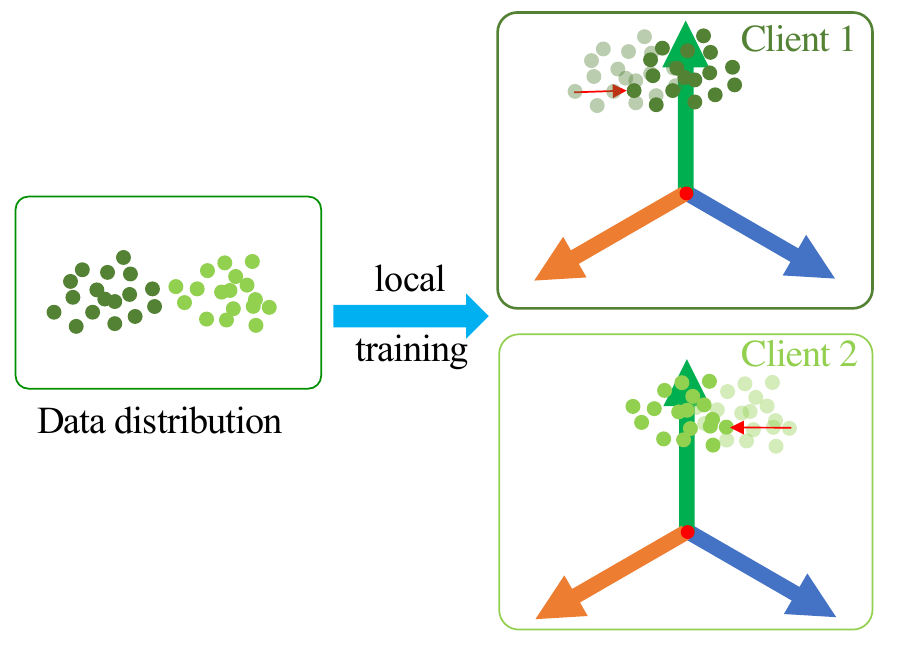}
    \caption{The phenomenon of non-iid that $\mathcal{P}_k(x|y)$ are different. We show the case that the samples of a client are not general. Specifically, client $1$ only observes the samples in dark green, and client $2$ only observes the samples in light green. However, they are learning in different directions.}
\label{Fig:pxy}
\end{figure}
\section{Preliminaries}
\label{Preliminaries}
\subsection{Federated Learning with Non-iid Clients Data}
We denote $\{(\boldsymbol{x_i},y_i)\}_{i=1}^N$ as the training set with $N$ samples, where $y_i\in \boldsymbol{C}=\{1,2,\dots,C\}$ and $C$ is the number of classes.
Suppose we have $K$ clients, and each client has its local data distribution $\mathcal{D}_k=\mathcal{P}_k(x,y)$.
The data between clients cannot be shared due to privacy.
Each client does standard classification using their local data $\mathcal{D}_k$ independently.
It first uses the feature extractor $F_{\boldsymbol{\theta}_{k}}(\cdot)$ with parameters $\boldsymbol{\theta}$ and the shared classifier with weights $\{\boldsymbol{w}_c\}_{c=1}^C$. 
Usually, the softmax operator normalizes the scores of each class and returns the probability $p_{i,c}$ of each category.
The cross-entropy loss of federated learning is used to calculate the local loss of each client, as shown in Equation~\ref{eq:CELoss},
\begin{equation}
\mathcal{L}_k=-\sum_{i=1}^{N}\sum_{c=1}^C\mathcal{I}\{y_i=c\}\log p_{i,c}
\label{eq:CELoss}
\end{equation}
\noindent where $\mathcal{I}\{\cdot\}$ is the indication function.
Federated learning aims to optimize the combination of local losses, as shown in Equation~\ref{eq:FLloss},
\begin{equation}
\min_{\psi=\{\boldsymbol{\theta}, \{\boldsymbol{w}_c\}_{c=1}^{C}\}} \qquad\sum_{k=1}^{K}p_k\mathcal{L}_k(\psi;\mathcal{D}_k).
\label{eq:FLloss}
\end{equation}
\noindent where $p_k$ is the weight of client $k$, satisfying $p_k\geq 0, \sum_{k=1}^K p_k=1$. \looseness=-1

In this paper, we focus on label distribution non-iid challenge as categorized in~\cite{ProblemFed}, i.e., the $\mathcal{P}_k(y)$ may vary significantly among clients, and even the $\mathcal{P}_k(\boldsymbol{x}|y)$ may differ.
There are many possible scenarios for non-iid situations. 
For example, some clients may only access a part of categories, which makes the classifier prefer visible categories, as shown in Figure~\ref{Fig:py}. 
Even if there are samples of a specific category, these samples may not be universal and are biased, as shown in Figure~\ref{Fig:pxy}.
It will cause the classifier of the server to swing back and forth when ensemble.

\subsection{Drawbacks of Standard FL Algorithms}
FedAvg~\cite{FedAvg} is the most standard federated learning method, and follow-up methods are improved based on it.
Still, the performance of FedAVG is comparable. 
FedAVG optimizes the global model through multiple rounds of local and global procedures. 
Specifically, at the beginning of each round, a subset of clients $\boldsymbol{S\subseteq K}$ is selected, where $\mathcal{K}$ is the set of all clients.
Then each selected client $k\in \boldsymbol{S}$ executes the following local update procedure in parallel.\looseness=-1

During the local client update procedure, the client initializes with the server's global parameters, i.e., $\boldsymbol{\theta}_k\leftarrow \boldsymbol{\theta}_{\text{global}}, \boldsymbol{w}_{c,k}\leftarrow \boldsymbol{w}_c,\forall c\in\mathcal{C}$. 
Then, the client learns the updated parameters on its local private training set $\{(\boldsymbol{x}_{i,k},y_{i,k})\}_{i=1}^{N_k}$. 
However, the global model contains the complete set of proxies $\{\boldsymbol{w}_c\}_{c=1}^C$, while the local client may only observe a partial set of the whole categories set.
We denote the missing categories set as $\boldsymbol{M}$, and the observed categories set as $\boldsymbol{O}$ ($\boldsymbol{M}\cup\boldsymbol{O}=\boldsymbol{C},\boldsymbol{M}\cap\boldsymbol{O}=\emptyset$).
As for the proxies of missing classes, i.e., $\{\boldsymbol{w}_c\}_{c\in \boldsymbol{M}}$, they are only regarded as negative features.
It not only forces the features of missing classes to negative features but also shifts the features of observed features.
The proxies of observed classes, $\{\boldsymbol{w}_c\}_{c\in \boldsymbol{O}}$, are only pushed away from the other observed feature regions, which reduces the constraint of features.
Another problem is that if the samples of category $c$ are biased in clients, then the feature is not general and shifts.\looseness=-1

During the global server update procedure, the server collects the updated parameters from the selected clients.
It then simply averages them as new global parameters, i.e., $\boldsymbol{\theta}_{\text{global}}\leftarrow\frac{1}{|\boldsymbol{S}|}\sum_{k=1}^{|\boldsymbol{S}|}\hat{\boldsymbol{\theta}}_{c,k}, \boldsymbol{w}_c\leftarrow\frac{1}{|\boldsymbol{S}|}\sum_{k=1}^{|\boldsymbol{S}|}\hat{\boldsymbol{w}}_{c,k}$, where $c\in\left[1, C\right]$.
When the server aggregates, the problem of inaccurate client features will influence the global classifier even more seriously.\looseness=-1

\subsection{Neural Collapse}
\cite{NeuralCollapse} observes that the features and weight vectors of the classifier will dually converge to a particular geometric structure under the balanced data distribution and calls this phenomenon Neural Collapse.
Neural Collapse has four manifestations:
\begin{enumerate}[itemsep=2pt,topsep=0pt,parsep=0pt]
\item The features will converge to their class means.
\item Under a balanced data distribution, the class mean vectors will collapse to a simplex equiangular tight frame (ETF) geometric structure.
\item The classifier weights will dually converge to the mean vectors of the corresponding class.
\item The decision of the classifier is based on the Euclidean distances of the feature vectors.
\end{enumerate}
An ETF is a tight frame whose absolute inner products are identical, i.e., $\forall_{i\ne j}, \boldsymbol{x}_i^T\boldsymbol{x}_j = -\frac{1}{N-1}$.
It has the largest possible equiangular separation of $N$ unit vectors in $\mathbb{R}^d$ that can maximize the between-class variability.
Therefore, a classifier that satisfies the simplex ETF geometric structure should be an optimal answer for classification.
So, we take a random simplex ETF geometric structure as the consistent optimization target. 
Then the inconsistent optimization target across clients will be solved.\looseness=-1
\begin{definition}
\label{def:ETF}
Let $\boldsymbol{X}$ be a $d \times N$ matrix whose columns are $\boldsymbol{x}_1, \dots, \boldsymbol{x}_N $. The matrix $\boldsymbol{X}$ is called an equiangular tight frame if it satisfies three conditions.
\begin{enumerate}[itemsep=2pt,topsep=0pt,parsep=0pt]
    \item Each column has a unit norm.
    \item The columns are equiangular.
    \item The matrix forms a tight frame.
\end{enumerate}
\end{definition}

\begin{algorithm}[tb]
   \caption{FedAvg\_ETF}
   \label{alg:FedETF}
\begin{algorithmic}
   \STATE {\bfseries Input:} training data $\{(\boldsymbol{x}_{i,k},y_{i,k})\}_{i=1}^{N_k}$, the number of categories $C$, the number of clients $K$, the number of round $R$, weight hyper-parameter $\alpha$, batch size $B_k$, the warm-up round $R_{\text{warm}}$.
   \STATE {\bfseries Initialize:} Weights of global feature extractor $\boldsymbol{\theta}^{0}_{\text{global}}$, Weights of global classifier $\{\boldsymbol{w}\}_{c=1}^C$ (as a random ETF geometric structure).
   \FOR{round $t$ = 1 {\bfseries to} $R$}
   \STATE $\boldsymbol{S}_{t} = $ random set of $K$ clients
   \FOR{client $k \in \boldsymbol{S}_{t}$ in parallel}
   \STATE $\boldsymbol{\theta}_k^t\leftarrow\boldsymbol{\theta}_{\text{global}}^{t-1}$
   \FOR{$b=1$ {\bfseries to} $N_k / B_k$}
   \FOR{$(\boldsymbol{x}_{i,k}, y_{i,k}) \in \boldsymbol{D}_{b}^k$ {in parallel}}
   \STATE $\boldsymbol{f}_{i,k}^{t} \leftarrow F_{\boldsymbol{\theta}_{k}^t}\left(\boldsymbol{x}_{i,k}\right)$
   \IF{$t\geq R_{\text{warm}}$}
   \STATE $\boldsymbol{h}_{i,k}^t \leftarrow \boldsymbol{f}_{i,k}^t + \alpha\cdot\boldsymbol{\mu}_{y_{i,k}}^{t-1}$
   \ELSE
   \STATE $\boldsymbol{h}_{i,k}^t \leftarrow \boldsymbol{f}_{i,k}^t$
   \ENDIF
   \STATE $\mathcal{L}_{i,k}^t = -\log\left(\frac{\exp\left(\boldsymbol{w}_{y_{i}}^{\top}\boldsymbol{h}_{i,k}^t\right)}{\sum\limits_{j=1}^{c}\exp\left(\boldsymbol{w}_{j}^{\top}\boldsymbol{h}_{i,k}^t\right)}\right)$
   \ENDFOR
   \STATE $\mathcal{L}_{b,k}^t = \sum_{(\boldsymbol{x}_{i,k}, \; y_{i,k}) \in \mathcal{D}_{b}^k} \mathcal{L}_{i,k}^t$
   \STATE $\boldsymbol{\theta}_{k}^{t} \leftarrow \boldsymbol{\theta}_{k}^{t} - \eta\cdot\frac{\partial \mathcal{L}_{b,k}^t}{\partial \boldsymbol{\theta}_{k}^{t}}$
   \ENDFOR
   \ENDFOR
   \FOR{$j=1$ {\bfseries to} $C$}
   \STATE $\boldsymbol{\mu}_{j}^{t} \leftarrow \frac{1}{\vert\boldsymbol{S}_{t}\vert}\sum_{k \in \boldsymbol{S}_{t}}\frac{1}{\vert\mathcal{D}_{k,j}\vert}\sum_{i \in \mathcal{D}_{k,j}} \boldsymbol{f}_{i,k}^{t}$
   \ENDFOR
   \STATE $\boldsymbol{\theta}^{t}_{\text{global}} \leftarrow \frac{1}{K}\sum_{k \in \mathcal{S}_{t}}\boldsymbol{\theta}_{k}^{t}$
   \ENDFOR
\end{algorithmic}
\end{algorithm}

\section{Methods}
Considering the non-iid challenge, the $\mathcal{P}_k(y)$ may vary significantly among clients (as shown in Figure~\ref{Fig:py}), and even the $\mathcal{P}_k(\boldsymbol{x}|y)$ may differ too (as shown in Figure~\ref{Fig:pxy}).
For the former situation, inspired by the neural collapse~\cite{NeuralCollapse}, we propose to force all clients to a pre-assigned optimum directly.
For the latter situation, we propose to add a global memory vector for each category, which remedies the parameter fluctuation caused by the bias of the intra-class conditional distribution among clients.
Under the two simple modules, the performance of the central server is significantly improved, and the variance is slight.
Our method can be briefly described as Algorithm~\ref{alg:FedETF}, where $f$ is a feature vector representation after the backbone, $u_j$ is a global characterization vector of category $j$ obtained by GMV, and $h$ is an accurate characterization feature vector after GMV applying on the local feature vector $f$.

\subsection{Frozen ETF Structure}
\begin{figure}[tbp]
\centering
\includegraphics[width=0.9\linewidth]{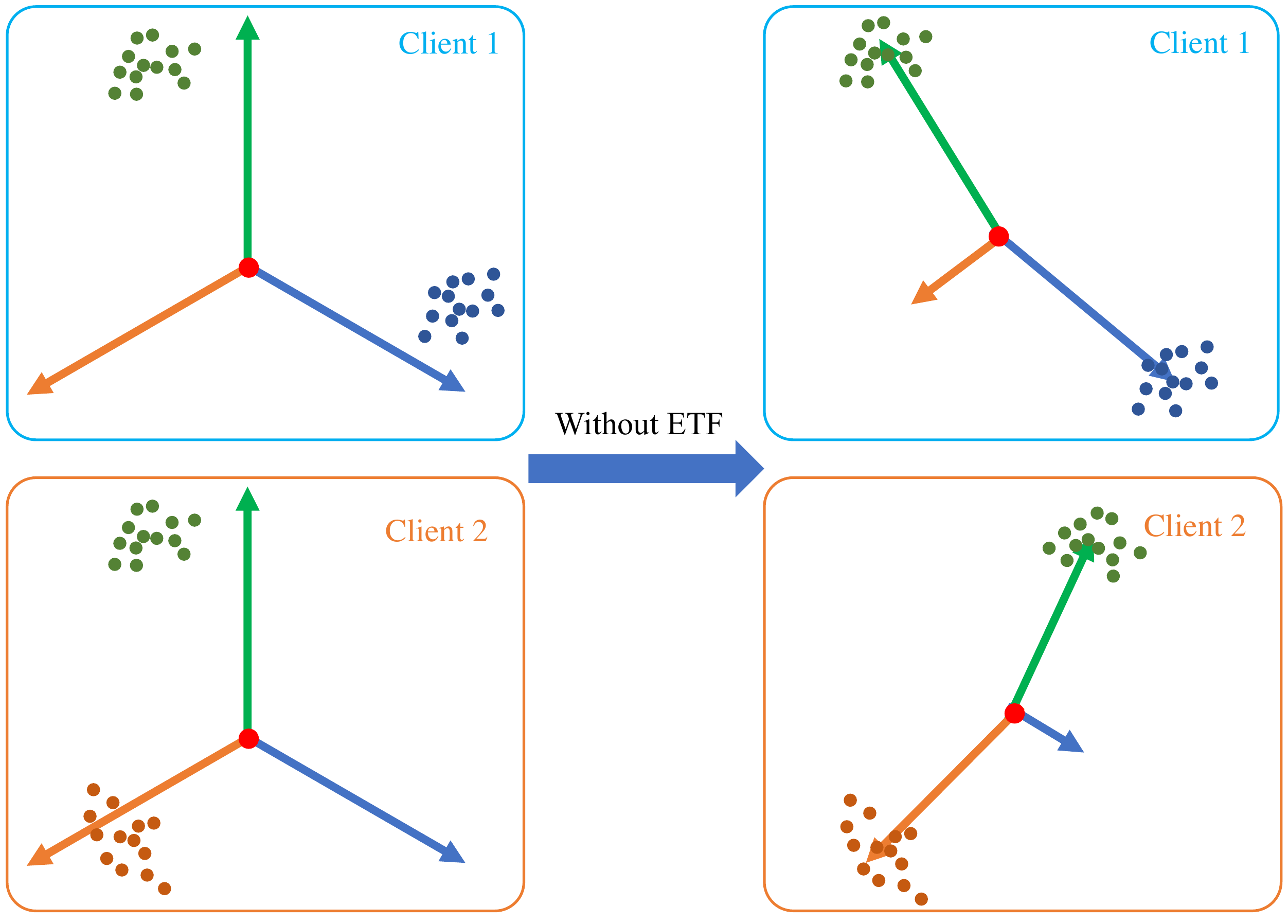}
\caption{Schematic diagram for using a random simplex ETF geometric structure to fix the classifier.}
\label{Fig:ETF}
\end{figure}
In this section, we consider one of the non-iid challenges that $\mathcal{P}_k(y)$ varies, as shown in Figure~\ref{Fig:py}. 
If the classifier is not constrained as in previous works, the feature vectors and the classifier weights will be learned in the local training. 
We assume that client $1$ can only observe green and blue categories, and client $2$ can only observe green and orange categories, as Figure~\ref{Fig:py} shows.
Then clients will converge to the optimal solution on their local dataset in the local training. 
That is, the weights of the two observed categories will be as far away as possible, and the unobserved classification weights will decrease. 
Thus, the model aggregated by the central server is affected.
So, we force each client to be optimized to a pre-assigned optimal target.

Although the non-iid situation in a client, the training set is balanced.
So, we can use the phenomenon of neural collapse that the classifier must finally converge to a simplex ETF geometric structure in a balanced distribution.
A simplex ETF describes a geometric structure that maximally separates the pair-wise angles of $C$ vectors in $\mathcal{R}^d, d \geq C - 1$, and all vectors have equal $\ell_2$-norms. 
When $d = C - 1$, the ETF is reduced to a regular simplex.
We produce the ETF geometric structure by the Equation~\ref{eq:ETF}
\begin{equation}
\boldsymbol{W}^{*} = \sqrt{\frac{c}{c - 1}} \boldsymbol{P}(\boldsymbol{I}{c} - \frac{1}{c}\boldsymbol{1}{c}\boldsymbol{1}{c}^{\mathrm{T}}),
\label{eq:ETF}
\end{equation}
\noindent where $\boldsymbol{P} \in \mathbb{R}^{d \times c}, (d \geq c)$ is a partial orthogonal matrix such that $\boldsymbol{P}^{\mathrm{T}}\boldsymbol{P} = \boldsymbol{I}{c}$, $\boldsymbol{I}{c}$ is the $c \times c$ identical matrix and $\boldsymbol{1}{c} \in \mathbb{R}^{c \times 1}$ is a vector filled with $1$. 
If $C=4$, then a 3D simplex ETF geometric structure is like the molecular structure of methane.
The ETF structure is so independent and balanced that it can be regarded as the optimum structure of a classifier.\looseness=-1

Specifically, we initialize and freeze the structure of the global classifier $\{\boldsymbol{w}_c\}_{c=1}^{C}$ as a random simplex ETF structure.
During the training phase, we fix the weights of the classifier $\boldsymbol{w}$, i.e., only the weights of backbone $\boldsymbol{\theta}$ are learnable.
In that way, each client trains under the constraints that there are $C$ independent categories with no bias and should be converged to the optimum.
As shown in Figure~\ref{Fig:ETF}, after initializing and fixing the structure of an ETF structure, the weights of the classifier will not damage, and only the points are moving, i.e., the features are learnable.

\begin{figure}[tbp]
\centering
\belowcaptionskip=-1pt
\includegraphics[width=0.4\linewidth]
{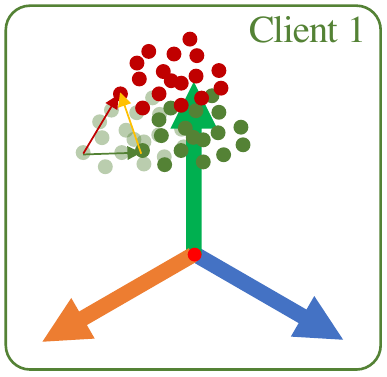}
\caption{Schematic diagram of global memory vectors (GMV). The \textcolor{yellow}{yellow arrow} indicates our designed GMV. \textcolor{green}{Green samples} show the results without GMV. \textcolor{red}{Red samples} show the results with GMV, and the direction of its learning will consider the global information about the category.}
\label{Fig:GMV}
\end{figure}

\subsection{Global Memory Vectors}
As the local samples are not general and the intra-class sub-distribution among clients is biased, there is fluctuation in the training phase.
Therefore, we add a global memory vector (GMV) for each category to remedy this fluctuation, as shown in Figure~\ref{Fig:GMV}. 
Specifically, we use $\boldsymbol{h}_{i,k}^t \leftarrow \boldsymbol{f}_{i,k}^t + \alpha\cdot\boldsymbol{\mu}_{y_{i,k}}^{t-1}$ instead of $\boldsymbol{h}_{i,k}^t \leftarrow \boldsymbol{f}_{i,k}^t$ , as shown in Algorithm~\ref{alg:FedETF}.
In this way, the local training will not only be affected by the local dataset but also be constrained by global memory vectors.
Like the training model, we spread global memory vectors to each client before local training and update global memory vectors by all clients during aggregation.
Since those feature representations are not good at the beginning of training, inaccurate global memory vectors will affect the local learning phase.
Therefore, we add global memory vectors after feature representations are learned well and initialize global memory vectors as the average weight of feature vectors.
Our global memory vectors can help both local training and global aggregation.

\begin{table}[t]
\caption{Performance comparisons with standard FL algorithms (VGG11). Columns correspond to three scenes. The number of local epochs is $2$. The average accuracy of the last $50$ rounds. $C$ means the Cifar dataset. The best performances are marked as bolded. FedAvg\_ETF means our method based on FedAvg.}
\label{tab:methods}
\vskip 0.15in
\begin{center}
\begin{small}
\begin{sc}
\resizebox{\linewidth}{!}{
\begin{tabular}{lccc}
\toprule
Method & C10-100-2 & C10-100-5 & C100-100-20 \\
\midrule
FedAvg & 69.6 & 82.0 & 49.8\\
FedMMD & 70.6 & 82.6 & 50.1\\
SCAFFOLD & 71.8 & 82.6 & 50.1\\
FedProx & 68.4 & 82.7 & 50.4\\
FedAwS & 71.5 & 82.8 & 49.7\\
FedNova & 70.0 & 81.1 & 47.4\\
FedOpt & 75.5 & 83.2 & 47.5\\
MOON & 69.5 & 83.1 & 48.0\\
FedRS & 73.9 & 83.4 & 47.5\\
SCAFFOLD/RS & 83.8 & 73.0 & 50.6\\
FedDyn & - & 82.5 & 51.0\\
\hline\\
FedAvg\_ETF & \textbf{79.8} & \textbf{84.6} & \textbf{51.3}\\
improvement & \textbf{+3.0} & \textbf{+1.2} & \textbf{+0.9}\\
\bottomrule
\end{tabular}}
\end{sc}
\end{small}
\end{center}
\vskip -0.1in
\end{table}

\section{Experiments}
\label{exp.}
\subsection{Basic Settings}
We then compare the performances based on Cifar10 / Cifar100. 
We construct label distribution shift scenes via label partitions, as done in several previous works~\cite{FedMD,FedPos,FedNoniid,FedRS}. 
Specifically, we decentralize the data onto $K=100$ clients, with each client only having a subset of classes. 
Moreover, as in previous works~\cite{FedRS}, we use VGG11 as the backbone and construct three scenes: Cifar10-100-5, Cifar10-100-2, and Cifar100-100-20.
Take Cifar10-100-5 as an example, we use the Cifar-10 dataset as a training set, and each client contains $5$ categories on average and $100$ samples for each category.

In each global round, we randomly select $10\%$ clients.
In this work, we do not care about the framework setting of federated learning, including the model structure of the server, the local training process, and the global model aggregation process.
We take $1000$ global rounds, a batch size of $64$, and a weight decay of $5e-4$. 
We use SGD with momentum $0.9$ as the optimizer and use a constant learning rate of $ 0.03$.
We report the accuracy of the aggregated model on the global test set, i.e., the test partition of Cifar10 / Cifar100. The default setting of ETF is $1.0$.\looseness=-1

\begin{table}[t]
\caption{Performance of using ETF geometric structure under various federated learning methods. All methods are evaluated on the Cifar dataset. The best performances are marked as bolded.}
\label{tab:plugin}
\vskip 0.15in
\begin{center}
\begin{small}
\begin{sc}
\resizebox{\linewidth}{!}{
\begin{tabular}{lccc}
\toprule
Method & C10-100-2 & C10-100-5 & C100-100-20 \\
\midrule
FedAvg & 69.6 & 82.0 & 49.8\\
+ETF & 71.4 & 82.8 & 48.7\\
improvement & \textbf{+1.8} & \textbf{+0.8} & -1.1\\
\hline\\
SCAFFOLD & 71.8 & 82.6 & 50.1\\
+ETF & 80.8 & 86.1 & 51.5\\
improvement & \textbf{+9.0} & \textbf{+3.5} & \textbf{+1.4}\\
\hline\\
FedNova & 70.0 & 81.1 & 47.4\\
+ETF & 71.5 & 81.1 & 47.9\\
improvement & \textbf{+1.5} & 0.0 & \textbf{0.5}\\
\hline\\
FedOpt & 75.5 & 83.2 & 47.5\\
+ETF & 76.4 & 83.3 & 48.7\\
improvement & \textbf{+0.9} & \textbf{+0.1} & \textbf{1.2}\\
\hline\\
MOON & 69.5 & 83.1 & 48.0\\
+ETF & 70.5 & 83.2 & 48.7\\
improvement & \textbf{+1.0} & \textbf{+0.1} & \textbf{+0.7}\\
\bottomrule
\end{tabular}}
\end{sc}
\end{small}
\end{center}
\vskip -0.1in
\end{table}

\begin{figure*}[tbp]
\centering
\subfigbottomskip=2pt
\subfigcapskip=-5pt
\subfigure[In Cifar10-100-2 with local epoch$=2$]{
    \includegraphics[width=0.48\linewidth]{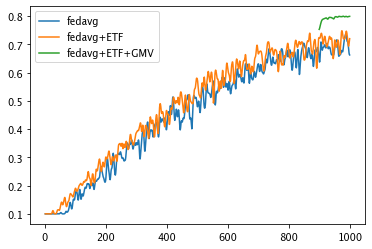}}
\subfigure[In Cifar10-100-2 with local epoch$=5$]{
    \includegraphics[width=0.48\linewidth]{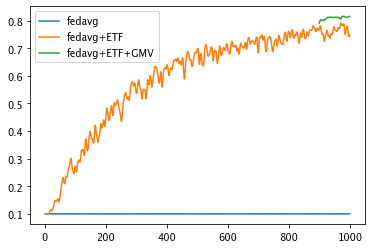}}\\
\subfigure[In Cifar10-100-5 with local epoch$=2$]{
    \includegraphics[width=0.48\linewidth]{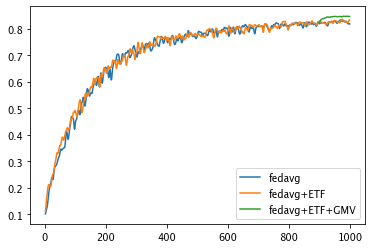}}
\subfigure[In Cifar10-100-5 with local epoch$=5$]{
    \includegraphics[width=0.48\linewidth]{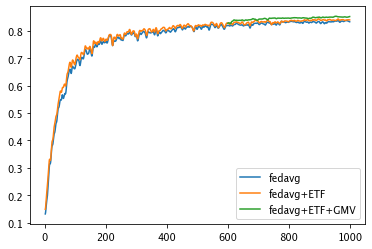}}
  \\
\subfigure[In Cifar100-100-20 with local epoch$=2$]{
    \includegraphics[width=0.48\linewidth]{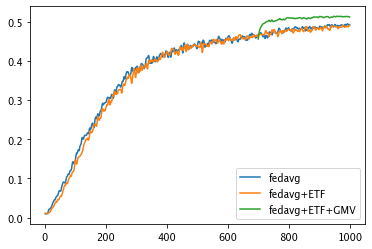}}
\subfigure[In Cifar100-100-20 with local epoch$=5$]{
    \includegraphics[width=0.48\linewidth]{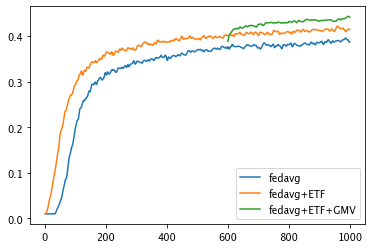}}
\caption{The relationship between round and accuracy. We report three scenes: Cifar10-100-2, Cifar10-100-5, Cifar100-100-20, and two values of the local epoch. Global memory vectors (GMV) add to training when the network is learned well. We only draw the results under the optimal setting, so they start at different epochs. 'Fedavg+ETF+GMV' is the same as our FedETF in Table~\ref{tab:methods}. We use a $\sigma=2$ gaussian kernel to smoother the curve.}
\label{Fig:round-acc}
\end{figure*}

\subsection{Quantitative Evaluation}
We compare our method with other FL methods, including the basic method FedAvg~\cite{FedAvg}; 
FedMMD~\cite{FedMMD}, FedProx~\cite{FedProx} and FedDyn~\cite{FedDyn} based on regularization; 
SCAFFOLD~\cite{scaffold} based on momentum and controllable variates; 
FedAwS~\cite{FedAwS} based on spreading out; 
FedNova~\cite{FedNova} based on adaptive local steps for each round;
MOON~\cite{Moon} based on contrastive learning;
FedOpt~\cite{FedOpt} based on adaptive optimization; 
and FedRS\cite{FedRS} based on restrict softmax.
As shown in Table~\ref{tab:methods}, our method surpass all the above methods only under the basic framework: FedAvg.
The best performances are marked as bolded.

We show the relationship between round and accuracy in Figure~
\ref{Fig:round-acc} for our method based on FedAvg.
From these three scenes and two local epochs, we find that ETF geometric structure can make variance smaller and have better accuracy.
Furthermore, global memory vectors further improve performance.
We only draw the round-accuracy curve for each configuration's optimal performance with global memory vectors. 
The hyper-parameters of global memory vectors in each figure are different, and the specific parameters are shown in Table~\ref{tab:alpha}.

\begin{table*}[t]
\caption{Ablation study on the weight of global memory vectors $\alpha$ and the number of warm-up rounds $R_{\text{warm}}$. The best performance is marked in bold, and the second is marked in underlining.}
\label{tab:alpha}
\vskip 0.15in
\begin{center}
\begin{small}
\begin{sc}
\begin{tabular}{cccccccc}
\toprule
\multirow{2}{*}{$\alpha$} & \multirow{2}{*}{Warm-up Round} & \multicolumn{3}{c}{2} & \multicolumn{3}{c}{5}\\ 
\cmidrule{3-8}
 & & C10-100-2 & C10-100-5 & C100-100-20 & C10-100-2 & C10-100-5 & C100-100-20 \\
\midrule
0 & - & 71.4 & 82.8 & 48.7 & 77.0 & 84.1 & 41.5\\
\hline
\multirow{5}{*}{0.1} & 500 & 71.9 & 83.4 & 50.9 & 76.3 & 84.1 & 42.3\\
& 600 & 73.0 & 83.7 & \textbf{51.3} & 77.6 & 84.3 & 42.5\\
& 700 & 74.4 & 84.0 & \underline{51.1} & 79.3 & 84.2 & 41.8\\
& 800 & 75.3 & 84.1 & 50.8 & \underline{81.2} & 84.2 & 42.1\\
& 900 & 75.7 & 84.1 & 50.4 & 81.0 & 83.5 & 40.8\\
\hline
\multirow{5}{*}{0.5} & 500 & 71.5 & 83.7 & 50.5 & 78.1 & 84.5 & 43.1\\
& 600 & 74.7 & 84.2 & 50.8 & 78.8 & 84.6 & 43.5\\
& 700 & 76.5 & \underline{84.4} & \textbf{51.3} & 79.5 & 84.6 & 43.1\\
& 800 & 78.0 & \underline{84.4} & 50.8 & 80.3 & 84.8 & 42.9\\
& 900 & \underline{79.5} & \textbf{84.6} & 50.6 & \textbf{81.3} & 84.6 & 42.3\\
\hline
\multirow{5}{*}{1.0} & 500 & 70.8 & 83.3 & 50.4 & 77.2 & \underline{85.0} & \underline{43.7}\\
& 600 & 74.7 & 83.9 & 50.4 & 78.8 & \textbf{85.2} & \textbf{43.8}\\
& 700 & 76.2 & 84.2 & 50.6 & 78.9 & 84.9 & 43.5\\
& 800 & 78.9 & 84.1 & 50.2 & 79.9 & 84.8 & 43.3\\
& 900 & \textbf{79.8} & 83.7 & 49.3 & 79.9 & 84.8 & 42.6\\
\bottomrule
\end{tabular}
\end{sc}
\end{small}
\end{center}
\vskip -0.1in
\end{table*}

We also find that our method can be easily combined with other methods, e.g., FedAvg~\cite{FedAvg}, SCAFFOLD~\cite{scaffold}, 
FedNova~\cite{FedNova}, FedOpt~\cite{FedOpt}, and MOON~\cite{Moon}, 
as shown in Table~\ref{tab:plugin}. 
GMV needs to adjust the corresponding optimal parameters under different settings. 
Since the previous methods have proposed multiple improvements for the non-iid problem, whose motivation is similar to GMV, we only verify the versatility of ETF, i.e., only fix the global classifier to a random ETF geometric structure.
After combing our designs, the performance can improve further.
The improvements are enormous in C10-100-2.
For FedAvg in C100-100-20, although adding ETF alone does not improve, the performance can be improved to $51.3$ after adding GMV. 
We think it is because the imbalance within the category seriously affects the ETF.

\begin{table}[t]
\caption{Ablation study on each component in FedAvg\_ETF. We do the experiments both in local epochs $2$ and $5$. The best performances are marked as bolded. $*$: When we reproduce FedAvg in local epoch$=10$ and scene$=$cifar-100-2 many times, we find the performance deficient, which is not reported in its paper.}
\label{tab:components}
\vskip 0.1in
\begin{center}
\begin{small}
\begin{sc}
\resizebox{\linewidth}{!}{
\begin{tabular}{clccc}
\toprule
Epoch & Method & C10-100-2 & C10-100-5 & C100-100-20 \\
\midrule
\multirow{3}{*}{2} & FedAvg & 69.6 & 82.0 & 49.8\\
& +ETF & 71.4 & 82.8 & 48.7\\
& +GMV & 77.4 & 82.7 & 50.3\\
& +ETF\&GMV & \textbf{79.8} & \textbf{84.6} & \textbf{51.3}\\
\hline\\
\multirow{3}{*}{5} & FedAvg & 10.0$^{*}$ & 83.6 & 39.0\\
& +ETF & 77.0 & 84.1 & 41.5\\
& + GMV & 10.0$^{*}$ & 83.8 & 37.2\\
& +ETF\&GMV & \textbf{81.3} & \textbf{85.2} & \textbf{43.8}\\
\bottomrule
\end{tabular}}
\end{sc}
\end{small}
\end{center}
\vskip -0.2in
\end{table}

\begin{table}[t]
\caption{Ablation study of Hyper-parameters in ETF geometric structure when the local epoch is set to 2. We do not use GMV in this experiment. The best performance is marked in bold, and the second is marked in underlining.}
\label{Tab: ETF}
\vskip 0.15in
\begin{center}
\begin{small}
\begin{sc}
\begin{tabular}{cccc}
\toprule
ETF & C10-100-2 & C10-100-5 & C100-100-20 \\
\midrule
None & 69.6 & 82.0 & \textbf{49.8}\\
0.5 & 68.6 & 82.7 & 48.4\\
1.0 & \textbf{71.4} & \underline{82.8} & \underline{48.7}\\
1.5 & \underline{71.0} & \textbf{83.0} & 47.2\\
\bottomrule
\end{tabular}
\end{sc}
\end{small}
\end{center}
\vskip -0.1in
\end{table}

\subsection{Ablation Studies}
\subsubsection{Validity of Each Component}
\label{sec:component}
To study the impact of each component of FedAvg\_ETF, we investigate them through extra experiments, as shown in \text{Table~\ref{tab:components}}. 
The results show that Both ETF geometric structure and global memory vectors of FedAvg\_ETF are effective. 
As components gradually increase, the final accuracy also increases accordingly.
We can see that the initial performance (acc) is boosted intensely under each setting, which is rather impressive.\looseness=-1

\subsubsection{Hyper-parameters for ETF geometric structure}
\label{sec:hyperETF}
We further show the influences of the hyper-parameter in a simplex ETF geometric structure.
The results are reported in \text{Table~\ref{Tab: ETF}}. 
Since each setting of ETF geometric structure has different best hyper-parameters of global memory vectors, we do not use global memory vectors in this experiment.
We achieve relatively better performance when $\beta=1.0$ in all three scenes.
However, we still suggest selecting the optimal parameters through experiments since the optimal target is different for different datasets and scenes.

\subsubsection{Hyper-parameters for Global Memory Vectors}
\label{sec:hyperGMV}
We investigate the impact brought by the weight of global memory vectors and the number of warm-up rounds.
The results are shown in \text{Table~\ref{tab:alpha}}.
The best settings are marked as bold.
In different scenes, the best settings are different.
But we find that global memory vectors should be added after the features are learned well. 
That is, the warm-up rounds should be large.
Moreover, global memory vectors are essential to the performances, so the weight of global memory vectors $\alpha$ should not be tiny.
We suggest to choose the hyper-parameter $R_{\text{warm}}$ from 500, and $\alpha \in \left[0.5, 1\right]$.

\subsection{Other Datasets}
We also verify our method in FeMNIST. 
FeMNIST is the Federated Extended of MNIST, which is derived from the Leaf~\cite{Leaf} repository.
There are 3550 users in FeMNIST, and each user owns 229 samples on average.
We still use FedAvg as the base algorithm and set the total number of clients $K$ to 100. 
Since FeMNIST is simpler than CIFAR, we use MLPNet as the backbone to avoid overfitting.
The results are shown in Table~\ref{tab:fmnist}.
\begin{table}[t]
\caption{Performance on FMNIST scene.}
\label{tab:fmnist}
\vskip 0.15in
\begin{center}
\begin{small}
\begin{sc}
\begin{tabular}{ccc}
\toprule
Local Epoch & Method & FeMNIST\\
\midrule
\multirow{2}{*}{2} & None & 62.0\\
& FedETF & \textbf{73.4}\\
& Improvement & +11.4\\
\hline
\multirow{2}{*}{5} & None & 78.9\\
& FedETF & \textbf{85.2}\\
& Improvement & +6.3\\
\bottomrule
\end{tabular}
\end{sc}
\end{small}
\end{center}
\vskip -0.1in
\end{table}

\section{Conclusion}
\label{Con.}
We study the non-iid problem in federated learning. 
Inspired by the phenomenon of neural collapse, we propose directly adding a consistent optimization target across local clients.
Specifically, we initialize the classifier to a random simple ETF geometric structure and fix it during training, only learning the weight of the feature extractor. 
Moreover, we propose adding a global memory vector for each category to remedy the parameter fluctuation caused by the bias of the intra-class condition distribution among clients after guaranteeing all clients converge to the global optimum. 
Abundant experimental studies verify the superiority of our method.\looseness=-1

\nocite{langley00}

\bibliography{example_paper}
\bibliographystyle{icml2023}



\end{document}